# Modeling of Pruning Techniques for Deep Neural Networks Simplification

Morteza Mousa Pasandi, Mohsen Hajabdollahi, Nader Karimi, Shadrokh Samavi

*Department of Electrical and Computer Engineering*
*Isfahan University of Technology Isfahan 84156-83111, Iran*
morteza.mpasandi@ec.iut.ac.ir, m.hajabdollahi@ec.iut.ac.ir, nader.karimi@cc.iut.ac.ir, samavi96@cc.iut.ac.ir

*Abstract*— **Convolutional Neural Networks (CNNs) suffer from different issues, such as computational complexity and the number of parameters. In recent years pruning techniques are employed to reduce the number of operations and model size in CNNs. Different pruning methods are proposed, which are based on pruning the connections, channels, and filters. Various techniques and tricks accompany pruning methods, and there is not a unifying framework to model all the pruning methods. In this paper pruning methods are investigated, and a general model which is contained the majority of pruning techniques is proposed. The advantages and disadvantages of the pruning methods can be identified, and all of them can be summarized under this model. The final goal of this model is to provide a general approach for all of the pruning methods with different structures and applications.**

## I. Introduction

Convolutional neural networks (CNNs) are structures with strong feature extraction capabilities. CNNs are used in a lot of applications such as image processing, pattern recognition, speech recognition, etc. [1]. Efficient CNN models are equipped with any number of layers to extract deep features from the raw input data. Many numbers of layers with a lot of parameters impose some problems for the implementation of the CNN structure. These problems are due to computational as well as memory operations [2]. Convolutional processes and transactions related to the feature maps make a significant part of the computational and memory operations, respectively.

There are many studies in which the problem of CNN complexity and its model size reduction are addressed. These works are mainly based on quantization, model compression, and pruning. Quantization can reduce the number of bits required for neural computations [3]. Also, quantization can be used to compress the set of network weights and significantly reduce memory consumption. Other approaches, to deal with the CNN complexity, are shrinking, factorizing, and compressing pre-trained networks [4].

Recently by pruning techniques, significant parts of the computational operations of CNNs could be reduced. Pruning methods reduce CNN complexities by removing unnecessary elements in their structures [5]. Pruning can be used in different levels of CNN to decrease the utilized parameters with no significant accuracy drop. All of the CNN structures have redundant parameters which can be removed. The redundancy is originated from the brain mechanism, which can recover its functional capability even in the existence of reasonable neural damage [6]. Suppose that we have a dataset $D = f(x; y)$, $i$ ranging from $1$ to $n$, and a given sparsity level $k$ (i.e., the number of in-important weights) pruning could be modeled as an optimization problem in Eq. 1 [7].

$$\min_w L(w; D) = \min_w \frac{1}{n} \sum_{i=1}^{n} l\left(w; (x_i, y_i)\right), \quad (1)$$
$$s.t. \quad w \in R^m \quad \|w\|_0 \leq k$$

Many research studies in the pruning area are aiming to propose new methods that enhance the results of the previous ones. However, there is a lack of a general model under which all the necessary steps for pruning are investigated. A general model can be used to better analysis of the previous methods and improve them. In this study, a general framework to model different techniques in the area of the pruning is proposed. By using this framework, it is possible to model the most studies conducted in the pruning of CNNs. To the best of our knowledge, there is not such a model in which all the pruning methods can be explored and investigated. The following advantages can be achieved using the proposed framework.

• A brief view of different techniques introduced in the pruning methods.

• Better analysis of the pruning methods to identify their advantages and disadvantages.

• Using all the proposed techniques to develop a new pruning method better.

In Fig. 1 the general model for pruning with all details is depicted. As illustrated in Fig. 1, in the proposed model, there are four main stages including, 1) the training model, 2) objective selection, 3) pruning, and 4) tuning. In the training stage, all parameters are adjusted with values that depend on the intended objectives of the network. After the training stage, objectives of pruning forms the next steps of pruning. In the pruning step, redundant parts of the network structures are removed to reduce the model complexity. Different methods and techniques are introduced for the pruning step, and selecting the best one is essential to reach the final pruning objectives. Pruning can be lead to an accuracy drop due to removing some model parts. However, this accuracy drop can be compensated by a fine-tuning method. After each tuning, the final model is evaluated in terms of different metrics, such as performance and accuracy. Evaluation in each step could vary during different pruning cycles. The pruning process is terminated when the relevant results are observed.

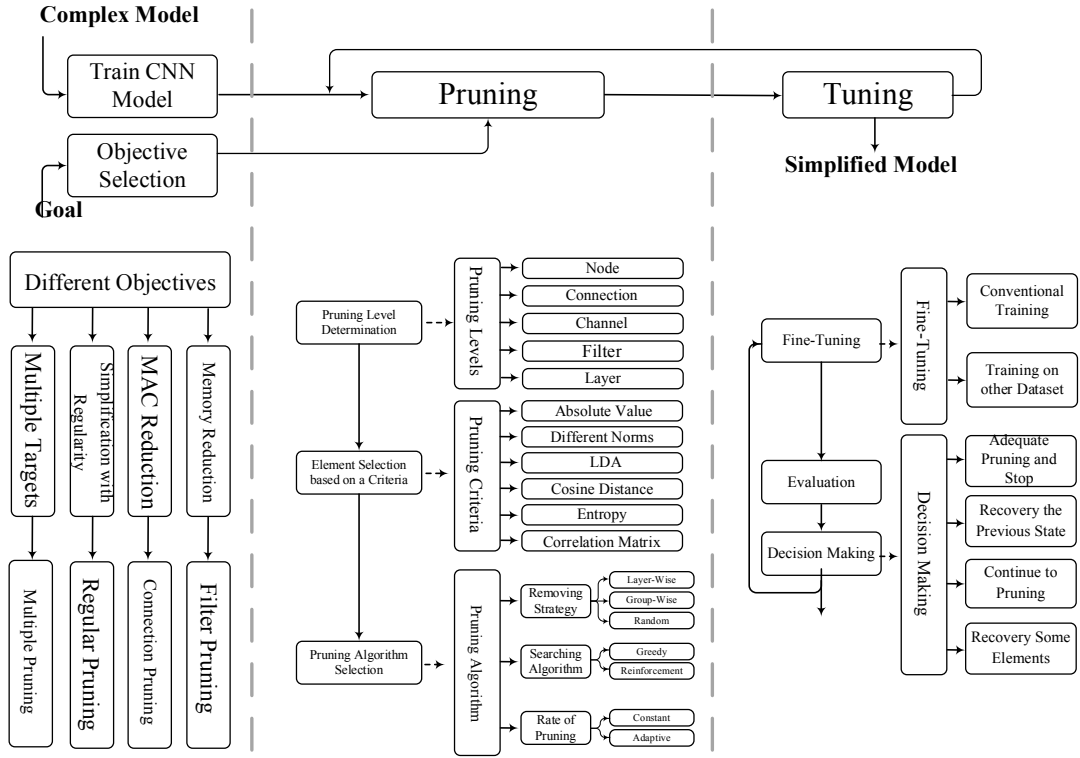

Fig 1. General framework for pruning.

In the next sections of this paper, various pruning methods for neural networks are summarized under the proposed model illustrated in Fig. 1. The remaining parts of this paper are organized as follows. In Section II, training and selection of different goals are presented. In Section III, different pruning methods are presented. In section IV, Fine-tuning and evaluation methods are summarized under the proposed model. Finally, Section V is dedicated to the discussion and conclusion

## II. Training Model and Objective Selection

To identify which variables and parameters in a CNN model are redundant, a training step is required. Training can proceed until the model reaches its final accuracy or can be in the form of a few epochs.

After training the base network structure, the objective of the pruning should be specified based on the problem goals. Different objectives for the pruning of the networks can be imagined which leads to the goal of simplicity. CNNs can be simplified in different views, which make different objectives. Different objectives are investigated in previous studies, which are shown in Fig. 2. Also, methods to realize the illustrated objectives are included in Fig. 2. In [5], pruning is used to reduce the number of parameters. The pruning method minimizes the number *multiply-and-accumulate* (MAC) operations; hence the objective could be the reduction of the MAC operations. Reducing the number of MAC operations is realized by reducing the number of connections.

In [8], pruning is utilized to reach a regular structure appropriate for the hardware implementation, which realized through an ordinary filter pruning. In [9], pruning is used to reduce the required memory consumption. In this work, feature maps as the primary energy-consuming part of CNNs are considered.

Pruning can be used with other objectives and even newly defined objectives that are not defined yet. Some of the newly established pruning objectives could be criteria such as speedup and run-time. In the case of such objectives, pruning is performed to reduce the critical path of the model.

It is worth mentioning that some previous studies have used multiple objectives to have a suitable result [10]. From a different perspective, pruning can be conducted in the case of convolutional layers, dense layers, and the whole model. In [5], only convolutional layers are pruned to reduce the computational power. In [11], dense layers are pruned to reduce the model's complexity. Also, in [8], the whole model is pruned to create a slimmer model with a reduced number of operations.

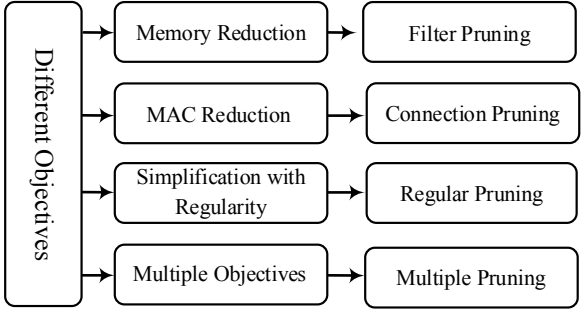

Fig. 2. Different objectives in the pruning methods

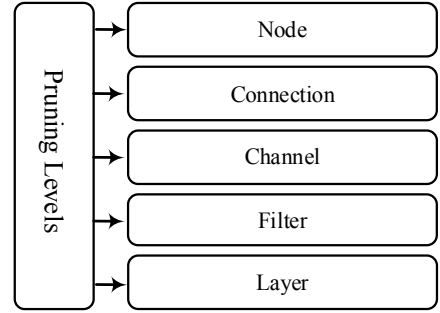

Fig. 4. Pruning levels

## III. PRUNING

In this section, we are going to explain the central part of a pruning algorithm. As illustrated in Fig. 3, pruning techniques have three aspects, which considering all of them, is essential to design an efficient pruning algorithm. These three steps are explained as follows.

*A. Pruning Level Determination*

As illustrated in Fig. 4, different elements can be regarded for pruning including node, connection, channel, filter, and layer. Previous studies concentrated on different elements for pruning, and some of them are based on pruning multiple elements. In [11] and [12], pruning is conducted on nodes, and in [13], network connections are pruned. In [14] redundant channels and in [15], redundant filters are pruned to make a simple structure. The level of pruning can be changed in any stage of pruning. It depends on the pruning objective and how the model can be changed.

*B. Element Selection for Pruning*

From different perspectives, pruning can be based on different criteria. Various criteria can determine redundant elements in the pruning process. In Fig. 5, these criteria are summarized. The absolute value of the connection weights [6] and L1-norm [5] are widely used as criteria of pruning. In [16] and [7] a set of rules for pruning are introduced, which perform better than those of previous works. In [12], pruning criteria are tested and compared in the case of medical image applications. In [17], a modified LDA is used for selecting the redundant filters. Luo et al., in [18], employed L1-norm and reconstruction error minimization for channel selection. Their work is compared with the method in [10], which uses absolute values of the weight, and the better result is reported. He et al., in [11], used entropy and input and output norm as well as SVD for node selection. Liu et al. [8] introduced a scaling factor and sparsity-induced penalty, as well as the L1 norm to select channels and nodes.

He et al., in [14], used Frobenius norm for pruning channels. In [4], magnitudes of the weights are used to generate a binary mask indicating pruning elements. Singh et al. [19] employed entropy and cosine similarity to define new criteria for pruning. Also, in [16], a cosine distance is used to select unimportant filters. In [20], the output norm of each layer is used to prune previous layer parameters. In [21], a pairwise correlation matrix is defined by modifying the form of the Hessian matrix to obtain a pruning binary mask.

*C. Pruning Algorithm*

After selecting suitable criteria for pruning, a method is required for removing the unimportant weights as a pruning method. As illustrated in Fig. 6, the pruning algorithm can be varied in three cases, including removing strategy, searching algorithm, and rate of pruning. The removing policy specifies how the elements are removed from the model. The order of the element removal, location, and granularities are considered as the pruning strategy. The granularity of the pruning method can be significant to have an efficient pruning method.

Layer wise pruning, group-wise pruning, random pruning, and multi-pass pruning are methods employed by the previous studies in the area of pruning strategy. Layer wise pruning was used in [5], [10], [18], [15], [11], [22], multi-pass scheme was used by [16] and [8] and random pruning was used in [23].

The algorithm for searching and removing of the un-necessary elements play an essential role for pruning. The greedy search algorithm is used in [2], [24] and [25] for selecting the objective efficiently. Also, in [15] and [26],

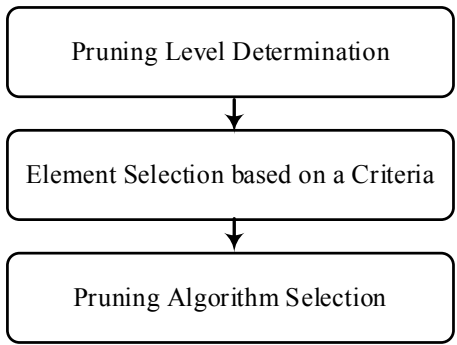

Fig. 3. Pruning Steps

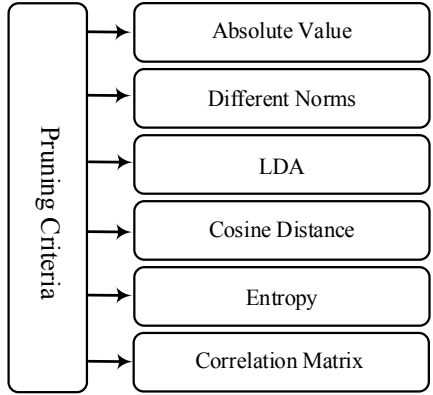

Fig. 5. Pruning criteria

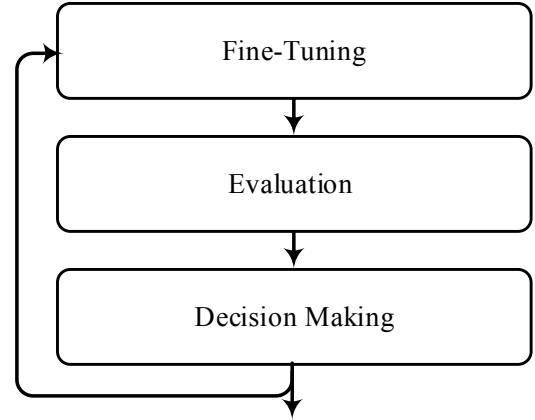

Fig. 7. Tuning steps

reinforcement learning is used for pruning the unnecessary filters.

Finally, an essential factor for the pruning algorithm can be considered the rate of pruning. A suitable rate of pruning allows the remaining elements in the model to recover the accuracy drop due to the pruning. Zhu et al. gradually pruned a network [4]. Also, pruning speed and its frequency can be changed during the pruning process.

## IV. TUNING

All steps in the tuning stage are illustrated in Fig.7. The tuning stage allows the network to recover its accuracy drop due to the pruning. After tuning, the performance of the network is evaluated, and based on this evaluation the next decisions are conducted. In Fig. 8, decision makings after pruning are presented. In the decision step, ways of avoiding network from unstable conditions are considered. Also, in the decision step the conditions in which the network pruning is terminated are considered. If an unstable station is observed, the previous state is recovered. Also, the instability can be prevented using advanced techniques by inserting the auxiliary elements to find ROI [27]. Hajabdollahi et al. use pruning methods for the segmentation of medical images [28-31]. In most previous studies, fine-tuning is conducted using conventional training algorithms in a few epochs. However, in some scenarios, fine-tuning is performed using another dataset to reach a better result. For example, Luo et al. in [18], tuned a network with Image-Net and tested on CUB-200 dataset. In [15], a novel decision part is included for pruning. In [9], the region in which accuracy is not suitable, are repaired to yield better accuracy. Shih et al. [27] employed the concatenation method to enhance accuracy degradation due to the pruning.

## V. UNIFICATION OF DIFFERENT PRUNING METHODS

In Table I, different pruning methods are summarized based on the model of Fig. 1, and are compared using the CIFAR10 image dataset. Thanks to the proposed general model, it is possible to represent any studies in Table I. As illustrated in Table I, various objectives are considered in different studies. L1-norm as the criteria and filter as the pruning level are widely used in different methods. Base on Table I, what makes pruning methods have different results is pruning principles and pruning algorithms. Also, in Table II, different pruning methods are represented under Fig 1 and compared in the case of the ImageNet image dataset. As Table II, in Table I, different objectives are considered in various studies, and the number of filters is widely used as the pruning criteria. In the case of the Image-Net dataset, significant variations are employed for the pruning algorithms. From Table II, it can be deduced that, L1-norm with a multi-pass algorithm for pruning yields better results.

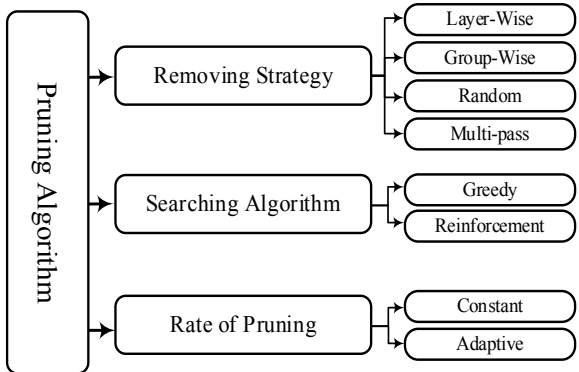

Fig. 6. Pruning algorithm

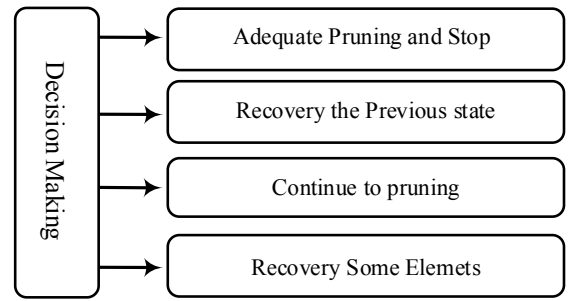

Fig. 8. Decision making

Table I. Summarization of different pruning methods based on the proposed general model (*Pr.* = pruning)

|  | *Reference [5]* | *Reference [12]* | *Reference [23]* |
|---|---|---|---|
| *Objective* | Computational cost | Multi Objective | Model Compression |
| *Pr. Level* | Filter | Filter | Whole(F/W) |
| *Pr. Criteria* | L1-norm | redundancy | L1-norm |
| *Pr. Algorithm* | Layer-wise Greedy | Reinforcement | CLASS-BLIND |
| *Fine Tuning* | 40 epochs | Yes | -- |
| *Decision Making* | Continue pruning | Try and Learn | Adequate pruning |
| *Accuracy / Pr. reduction* | 93.04 (VGG16) / 64/34<br>93.06 (Res56) / 13/27<br>93.3 (Res110) / 32/38 | 92.65 (VGG16)/ 83/45<br>90.62 (Res18) / 78/76 | 93.41 (VGG16) / 86/-<br>93.56 (Res110) / 53/- |
| *Dataset* | CIFAR10 | CIFAR10 | CIFAR10 |
|  | *Reference [8]* | *Reference [11]* | *Reference [13]* |
| *Objective* | Sparsity & Slim model | speed approach | Model Compression |
| *Pr. Level* | Whole (C/W) | Channel | Filter |
| *Pr. Criteria* | L1-norm | Frobenius norm | Geometric Median |
| *Pr. Algorithm* | Sparsity-induced Penalty | Layer wise | Multiple-Branch |
| *Fine Tuning* | Yes | 10 epochs | Yes and No |
| *Decision Making* | Continue pruning | Continue pruning | Continue pruning |
| *Accuracy / Pr. reduction* | 93.8 (VGG16)/ 88.5/51<br>93.56 (Res164) / 35/44 | 92.04 (Res56) / 70/- | 91.99 (Res20-n)/ 40/54<br>92.93 (Res32-n) / 40/52<br>93.49 (Res56-n)/ 40/52 |
| *Dataset* | CIFAR10 | CIFAR10 | CIFAR10 |

## VI. CONCLUSION

A new model for pruning was proposed in which the pruning studies were summarized. It was observed that in the pruning process, many techniques and procedures should be considered during the design of a new method for pruning. Under the proposed model, different approaches were investigated and summarized. Thanks to the proposed model for different pruning methods, a better analysis could be performed, and better insight into the state of the art methods is now possible. This study is the first work that is aiming to model different pruning methods with details.

Table II. Summarization of different pruning methods based on the proposed general model (*Pr.* = pruning)

|  | *Reference [25]* | *Reference [9]* | *Reference [13]* |
|---|---|---|---|
| *Objective* | Model Slimming | Time Complexity | Model Compression |
| *Pr. Level* | Whole model | Filter | Filter |
| *Pr. Criteria* | Normalization and Computation Graph | Better Accuracy | Geometric Median |
| *Pr. Algorithm* | Greedy Multi pass | Greedy Multi-path | greedy multi path |
| *Fine Tuning* | Yes | Yes | Yes and No |
| *Decision Making* | Continue pruning | Recover the previous states | Continue pruning |
| *Accuracy / Pr. reduction* | 65 (Res50)/ 92/93 | 92.28 (Res50) /56.73<br>93.07 (Res110) / 50.19<br>93.02 (Res152) / 74.69 | 91.08 (Res34)/ 30/41<br>92.32 (Res50)/ 40/53<br>93.56 (Res110)/ 40/42 |
| *Dataset* | Image-Net | Image-Net | Image-Net |
|  | *Reference [2]* | *Reference [26]* | *Reference [24]* |
| *Objective* | Network Acceleration | Computational Cost | Model Compression |
| *Pr. Level* | Filter | Channel | Channel |
| *Pr. Criteria* | L2 Norm | channel saliency and redundancy | L2 Norm |
| *Pr. Algorithm* | Layer by Layer greedy | Random ,reinforced and Adaptive | Greedy and Adaptive |
| *Fine Tuning* | Yes and No | No | Yes |
| *Decision Making* | Recover some elements | Continue Pruning | Adequate Pruning |
| *Accuracy / Pr. reduction* | 91.62 (Res34n)/ 30/41.5<br>92.87 (Res50y)/ 30/41.8 | 92.1 (Res50) / 40 | 91.13 (Res50)/70<br>85.87 (Res18)/70 |
| *Dataset* | Image-Net | Image-Net | Image-Net |